\documentclass[11pt,a4paper]{article}
\usepackage{emnlp2021}

\usepackage{times}
\usepackage{latexsym}
\usepackage{booktabs}
\usepackage{graphicx}
\usepackage{pgfplots}
\usepackage{subfig}
\usepackage[ruled,vlined]{algorithm2e}
\usepackage[normalem]{ulem}
\usepackage[T1]{fontenc}
\usepackage{multirow}
\usepackage{rotating}

\usepackage{microtype}

\usepackage{hologo}
\DeclareRobustCommand{\indicWords}[1]{%
  \raisebox{-\dp\strutbox}{%
    \includegraphics[page=\csname indicWords#1\endcsname]{indicWords}%
  }%
}

\title{Role of Language Relatedness in Multilingual Fine-tuning of Language Models: A Case Study in Indo-Aryan Languages}

 \author{Tejas Indulal Dhamecha\thanks{\ \ Equal contribution}, Rudra Murthy V$^*$, Samarth Bharadwaj,\\
 \bf Karthik Sankaranarayanan \\
        IBM Research, India\\
        \texttt{\{tidhamecha,rmurthyv,samarth.b,karsank\}@in.ibm.com}
        \AND
        \bf Pushpak Bhattacharyya\\
        Department of Computer Science and Engineering, IIT Bombay, India\\
        \texttt{pb@cse.iitb.ac.in}
        }
\date{}
\pgfplotsset{compat=1.16}
\usepackage[subtle]{savetrees}
\begin{document}
\maketitle
\begin{abstract}
We explore the impact of leveraging the relatedness of languages that belong to the same family in NLP models using multilingual fine-tuning. We hypothesize and validate that multilingual fine-tuning of pre-trained language models can yield better performance on downstream NLP applications, compared to models fine-tuned on individual languages. A first of its kind detailed study is presented to track performance change as languages are added to a base language in a graded and greedy (in the sense of best boost of performance) manner; which reveals that careful selection of subset of related languages can significantly improve performance than utilizing all related languages. The Indo-Aryan (IA) language family is chosen for the study, the exact languages being Bengali, Gujarati, Hindi, Marathi, Oriya, Punjabi and Urdu. The script barrier is crossed by simple rule-based transliteration of the text of all languages to Devanagari.  Experiments are performed on mBERT, IndicBERT, MuRIL and two  RoBERTa-based LMs, the last two being pre-trained by us. Low resource languages, such as Oriya and Punjabi, are found to be the largest beneficiaries of multilingual fine-tuning.  Textual Entailment, Entity Classification, Section Title Prediction, tasks of IndicGLUE and POS tagging form our test bed. Compared to monolingual fine tuning we get relative performance improvement of up to 150\% in the downstream tasks. 
The surprise take-away is that for any language there is a particular combination of other languages which yields the best performance, and any additional language is in fact detrimental.
\end{abstract}

\section{Introduction\footnote{\url{https://github.com/IBM/indo-aryan-language-family-model} % maybe even better to attach them as supple
}}

Transformer-based \cite{vaswani2017attention} language models (LM) have been proven to be extremely useful in variety of natural language processing (NLP) tasks. Some of the most notable models are GPT~\cite{radford2018improving}, GPT-2~\cite{radford2019language}, GPT-3~\cite{brown2020language}, BERT~\cite{devlin2019bert}, RoBERTa~\cite{liu2019roberta}, XLNet~\cite{yang2019xlnet}, and BART~\cite{lewis-etal-2020-bart}. To fine-tune a pre-trained language model for downstream tasks has become a \textit{de facto} approach in recent literature. 

We empirically study whether (and to what extent) do the related languages accentuate the performance of models in downstream tasks with multilingual fine-tuning\footnote{Fine-tuning a pre-trained model with a downstream task's training data for multiple languages.} in comparison to monolingual fine-tuning. To understand the quantitative advantage of including languages gradually, we explore the gradation of multilinguality by incrementally adding in new languages added one-by-one building up to an all-language multilingual fine-tuning. 

A good approximation for \textit{language relatedness} is their membership to the same \textit{language family} as languages of a family often share properties such as grammar, vocabulary, etymology, and writing systems. 
We choose the Indo-Aryan (IA) family for the study, since constituent languages 1) include low-resource languages, 2) have similar Abugida writing system,  3) are relatively understudied, and 4) are covered in a well-defined NLP benchmark IndicGLUE \cite{kakwani-etal-2020-indicnlpsuite}.
Further, the fact that all constituent languages except one use similar Abugida writing systems (rooted in the ancient Brahmi Script\footnote{\url{https://en.wikipedia.org/wiki/Brahmic_scripts}}) presents an opportunity for script normalization via transliteration.

Overall, although the general notion of language relatedness is explored, and the multilingual fine-tuning is explored in literature, the detailed linguistic understanding of role of language relatedness in multilingual fine-tuning remains understudied; even more so for IA family. Further, in this context the script-conversion aspect is not explored for multilingual fine-tuning.

To summarize, in this paper we seek to answer the following research questions (employing the Indo-Aryan language family as the experimental test-bed).

\begin{itemize}
    \item \textbf{RQ1:}  Does multilingual fine-tuning with a set of related languages yield improvements over monolingual fine-tuning (FT) on downstream tasks?  
    \item \textbf{RQ2:} Starting from monolingual FT, as each related language is gradually added for multilingual FT, to ultimately a multilingual FT with all related languages, how does the performance vary? In other words, should one use all related languages' data or only a sub-set of languages' data?
\end{itemize}

These inquiries are critical to understanding the right balance between per-language fine-tuning and massively multilingual fine-tuning, as the viable way forward. Additionally, we also explore the role of common script representation in multilingual FT of related languages.

To facilitate these inquiries, we utilize existing pre-trained models, namely IndicBERT, mBERT, and MuRIL, and also pre-train two language models for Indo-Aryan language family from scratch. We utilize various tasks of IndicGLUE \cite{kakwani-etal-2020-indicnlpsuite} as our test-beds. 

\section{Related Work}

Multilinguality aspect has been explored in context of pre-training language models, for effective transfer from one language to other, and in multilingual fine-tuning, to an extent.

\subsection{Multilingual Pre-training}

Multilingual LMs have enabled effective task fine-tuning across various languages. Notable examples include, multilingual BERT (mBERT)\footnote{\url{https://github.com/google-research/bert/blob/master/multilingual.md}}  model trained with 104 languages, and XLM \cite{lample2019cross} and XLM-RoBERTa \cite{conneau-etal-2020-unsupervised} trained with 100 languages.

In the context of Indic languages, three recent works on multilingual LMs have been IndicBERT, MuRIL, and Indic-Transformers language models.
IndicBERT \cite{kakwani-etal-2020-indicnlpsuite} focuses on languages belonging to the Indo-Aryan and Dravidian language families along with English. Along with 17 Indic languages and English, Multilingual Representations for Indian Languages (MuRIL)\footnote{\url{https://tfhub.dev/google/MuRIL/1}}\cite{khanuja2021muril}, utilizes English-Indic languages parallel corpus and the Roman transliterated counterparts to train an mBERT model. Similarly, Indic-Transformers \cite{jain2020indic} presents monolingual LMs for Hindi, Bengali, and Telugu. Recently, various types of word-embeddings are also trained for each language \cite{kumar-etal-2020-passage, kakwani-etal-2020-indicnlpsuite}.

These approaches focus on multilingual \textit{pre-training} of models. 
This means that once a multilingual LM is pre-trained, it is fine-tuned per task separately for each language. 

\subsection{Language Transfer}
It is understood that a multilingual model gains cross-lingual understanding from sharing of layers that allows the alignment of representations among languages; to the extent that large overlap of the vocabulary between the languages is not required to bridge the alignment \citep{conneau-etal-2020-emerging,DBLP:journals/corr/abs-1910-04708}. This property facilitates, zero-shot transfer between two related languages (e.g. Hindi and Urdu) reasonably well \cite{pires-etal-2019-multilingual}.
Performance for zero-shot transfer further improves when multilingual model is further aligned by utilizing parallel word or sentence resources \cite{kulshreshtha2020cross}.
Usually, the low-resource language members in a multilingual LM benefit by presence of related languages \cite{liu2020multilingual}.
Further, it is likely that presence of unrelated languages do not aid the multilingual training, but rather may lead to negative interference rooted in conflicting gradients \cite{wang-etal-2020-negative} or yield substantially poor transfer between unrelated languages (e.g. English and Japanese) \cite{pires-etal-2019-multilingual}. 
A recent work by \citet{dolicki2021analysing} focuses on establishing the connection between the effectiveness of zero-shot transfer and the linguistic feature of source and target languages; interestingly, they observe that the effectiveness of zero-shot transfer is a function of downstream task, in addition to the languages themselves.

The general understanding has been that language-specific FT serves as skyline, and, in these set of works, pursuit has been to get zero-shot transfer from related languages(s) closer to the skyline \citep{wu-dredze-2019-beto}.

\subsection {Multilingual Fine-tuning}
\citet{tsai-etal-2019-small} perform multilingual fine-tuning of 48 languages for downstream tasks of POS tagging and morphological tagging, and find these multilingual models to be slightly poorer compared to monolingual models. 
For morphological tagging and lemmatization  tasks, \citet{kondratyuk-2019-cross} makes similar observation regarding poor performance for the model fine-tuned with 66 languages in multilingual setting compared to monolingual fine-tuning (although, a second stage of per-language fine-tuning yields superior performance). 
\textit{These findings indicate that arbitrary collection of languages may not be suitable for improving downstream task performance; and that, a principled approach for selecting a set of languages may be preferable for multilingual fine-tuning.}  To this end, we hypothesize that language relatedness should be an important aspect to consider while selecting a language set for multilingual fine-tuning.

\citet{pires-etal-2019-multilingual} briefly explore language set selection based on topological features (syntactic word order).  \citet{wang-etal-2020-negative} explores multilingual fine-tuning in strictly bilingual settings.
Taking the language relatedness in considerations, \citet{tran-bisazza-2019-zero} show that joint fine-tuning with four European language is better than fine-tuning with only English in the specific task of universal dependency parsing. Unfortunately, it doesn't provide comparison with monolingual fine-tuning of all constituent languages.

We observe that there is a void regarding the systematic analysis to understand how a presence of related languages in the multilingual fine-tuning affects the performance on the target language.

\section{Methodology}

Traditionally, a pre-trained LM (such as mBERT) is used as base model, which is fine-tuned for a downstream task for a specific language (monolingual). 
In this work, we aim to evaluate the role of script and language relatedness in multilingual fine-tuning by employing the Indo-Aryan language family. Therefore, we include the following components to the approach: (1) multilingual fine-tuning, (2) transliteration, and (3) language models.

Next, we discuss these in detail.

\subsection{Multilingual Fine-Tuning}
As opposed to traditional monolingual fine-tuning for a downstream task, in multilingual fine-tuning the LM is fine-tuned once per task with the aggregate labelled corpus across languages. Intuitively, related languages should assist each other for a downstream task. To draw a parallel, a polyglot person (akin to a multilingual LM) good at guessing titles of passages written in one language can easily adapt this skill on another, albeit related, language with few exemplars. 
Arguably, when put together, a greater understanding of the downstream task arises compared to what each language would yield individually, and the relatedness of associated languages would play a key role in deciding the benefits of this approach. 

Therefore, to study this systematically, for a downstream task that is relevant for a variety of languages (e.g. part-of-speech tagging or named entity recognition), first the training sets of all languages are combined to create the multilingual task training set. Then, the base LM is fine-tuned on the multilingual task training corpus. This multilingual fine-tuning now yields a model per task, and not per task-language pair.

\subsection{Script Similarity and Transliteration}
Languages of a language family often use similar writing systems. For example, in the IA family, on one hand, Hindi, Bhojpuri, Magahi, Marathi, Sanskrit, and Nepali are written in Devanagari script. On the other hand, Bengali, Gujarati, Punjabi, and Oriya are written in their respective scripts. 
system of writing. 
As Indic languages have high lexical similarity \citep{bhattacharyya-etal-2016-statistical}, having a universal script for all these languages allows for model to exploit cross-lingual similarities. For example, the verb term for ``to go" is similar in Hindi (\indicWords{jaana} \verb jaanaa ), Urdu (\indicWords{jaanaU} \verb jaanaa ), Gujarati (\indicWords{jaavum} \verb javum ), Punjabi (\indicWords{jaanaP} \verb jaanaa ), Marathi (\indicWords{jaane} \verb jaane ), and Oriya (\indicWords{jibaku} \verb jibaku ), and Bengali (\indicWords{jao} \verb jao) with each language morphing it in different manners. We use indic-nlp-library \cite{kunchukuttan-etal-2015-brahmi} (for all but Urdu) and indic-trans tools \cite{Bhat:2014:ISS:2824864.2824872} (for Urdu) for transliteration to Devanagari.

\subsection{Language Models\label{sec:lm}}
For this study, we use the mBERT, IndicBERT, and MuRIL as existing pre-trained language models. Additionally, we pre-train language models (from scratch) specifically for only the Indo-Aryan languages, as other LMs contains languages of other families too.

\noindent\textbf{Pre-training Language Model From Scratch:}
We choose to pre-train RoBERTa \cite{liu2019roberta} transformer-based model as it has been shown to improve over BERT \cite{devlin2019bert} recently.  Existing pre-trained language models are trained on original script data. 
For fair study of effectiveness of transliteration, we wish to pre-train separate language models on original and transliterated corpus from scratch.
Our experimentation around transliteration makes existing pre-trained models (mBERT, IndicBERT, and MuRIL) somewhat incompatible. 
In other words, it would be akin to fine-tuning for an unseen language, albeit in a previously seen script. 
Thus, we settle upon pre-training contextual LM from scratch for this purpose.
Specifically, we train two LMs from scratch, one preserving the original scripts of corpora (IndoAryan-Original) and other after transliterating all corpora to Devanagari script (IndoAryan-Transliterated).

\section{Experimental Setup}
In this section, we describe the datasets used in our experiments, their pre-processing, and implementation details.

\subsection{Data}
To train the language models, we obtained text data from various sources including: Wikipedia Dump\footnote{\url{https://dumps.wikimedia.org/}}, WMT Common Crawl \footnote{\url{http://data.statmt.org/ngrams/raw/}}, WMT News CommonCrawl\footnote{\url{http://data.statmt.org/news-crawl/}}, Urdu Charles University Corpus \cite{11858/00-097C-0000-0023-625F-0,11858/00-097C-0000-0023-65A9-5}, IIT Bombay Hindi Monolingual Corpus \cite{kunchukuttan-etal-2018-iit}, Bhojpuri Monolingual Corpus \cite{kumar2018automatic}, and Magahi Monolingual Corpus\footnote{\url{https://github.com/kmi-linguistics/magahi}}. Various statistics of the collected corpus are reported in Table \ref{tab:lg-size}. Note the major imbalance in the data with Hindi being undoubtedly a high-resource language and likes of Magahi, Punjabi, and Oriya being low-resource languages.
The challenges of the data imbalance and insufficiency of data to train monolingual models for many of these languages is apparent from the statistics.

\begin{table}[]
\resizebox{0.45\textwidth}{!}{
\Large
\begin{tabular}{@{}lrrr@{}}
\toprule
\multirow{ 2}{*}{\textbf{Language}} & \multicolumn{1}{c}{\multirow{ 2}{*}{\textbf{\# Sentences}}} & \multicolumn{2}{c}{\textbf{\# Tokens}} \\\cline{3-4}

 &  & \multicolumn{1}{c}{\textbf{\# Total}} & \multicolumn{1}{c}{\textbf{\# Unique}} \\
\midrule
Hindi (hi) & 1552.89 & 20,098.73 & 25.01 \\
Bengali (bn) & 353.44 & 4,021.30 & 6.5 \\
Sanskrit (sa) & 165.35 & 1,381.04 & 11.13 \\
Urdu (ur) & 153.27 & 2,465.48 & 4.61 \\
Marathi (mr) & 132.93 & 1,752.43 & 4.92 \\
Gujarati (gu) & 131.22 & 1,565.08 & 4.73 \\
Nepali (ne) & 84.21 & 1,139.54 & 3.43 \\
Punjabi (pa) & 68.02 & 945.68 & 2.00 \\
Oriya (or) & 17.88 & 274.99 & 1.10 \\
Bhojpuri (bh) & 10.25 & 134.37 & 1.13 \\
Magahi (mag) & 0.36 & 3.47 & 0.15 \\
\bottomrule
\end{tabular}
}
\caption{ Statistics (in Millions) of monolingual corpora used in pre-training IndoAryan LMs from scratch.\label{tab:lg-size}}

\end{table}

\subsection{Data~Preparation~and~Implementation Details\label{sec:impl}}

As the first step, sentences are segmented from the text corpora. Then script converted version of the datasets is obtained by transliterating Bengali, Gujarati, Punjabi, Oriya, and, Urdu into Devanagari script. 
We additionally perform de-duplication to remove repeated sentences. The statistics of the resulting set are reported in Table \ref{tab:lg-size}.
We identify following two challenges that can affect pre-training negatively: 
1) data imbalance and 2) compute requirements.

\ \\
\noindent 1.~\textbf{Data imbalance:} As reported in Table \ref{tab:lg-size} the size of each language corpora differs up to four orders of magnitude, e.g. Hindi has 1552M sentences vs 0.36M in Magahi and 17.88M in Oriya.  The language bias can creep into the tokenizer and the language model pre-training, if left unattended. To mitigate bias in tokenizer training, we utilize a re-sampling strategy to reduce the data skew \cite{lample2019cross}.  
Specifically, samples are drawn following multinomial distribution with \emph{adjusted} probabilities. Adjusted probabilities are computed as $q_i = p_i^\alpha / \sum_j p_j^\alpha$ where $p_i = n_i/\sum_i n_i $, and $n_i$ is the number of samples in $i^{th}$ language. Before rescaling, the language distribution is in the range of 0.01\%-58\%, which changes to 5-12\% afterwards.

\ \\
% \item 
\noindent2.~\textbf{Compute requirements:} Depending on the computing infrastructure, running one training epoch can typically take few hundreds to (single digit) few thousands of GPU hours. To mitigate this, we utilize a variant of sharding technique outlined in Algorithm \ref{algo:trick} to pre-train the model in infrastructure with limited memory (<50GB) and compute (one v100 GPU). It depends on dividing each language corpus into manageable (into memory) chunks, termed as \emph{blocks}.
Each LM is trained over $\sim$50 sequential executions of Algorithm \ref{algo:trick} on a single v100 GPU machine and each execution running for a day, consuming about 1200 hours overall for pre-training.

\begin{algorithm}[!t]
\SetAlgoLined
\KwResult{Trained LM checkpoint}
   randomly select working language set\;
 \For{each language in working set}{
    randomly select least number of blocks containing x\% sentence of the language\;
    \For{each block}{
    \If{block is not cached}{
   tokenize the block\;
   persist on to storage\;
   }
   read tokenized block into dataset\;
    }
    }
\eIf{checkpoint found}{
   initialize model with checkpoint\;
   }
   {
   randomly initialize model\;
   }
 train LM to minimize MLM loss on balanced mini-batches\;
 \caption{Economical LM training for language family\label{algo:trick}}
\end{algorithm}

In the re-sampling step, exponent $s=0.1$ and scaling parameter of $\gamma=100$ are used. 
Byte-level BPE tokenizer \cite{radford2019language, wang2020neural} is used with vocabulary size of 110K. Trained LMs use 12 layers, 12 attention heads, hidden size of 768, and dropout ratio for the attention probabilities of 0.1. Our implementation uses Huggingface \cite{wolf-etal-2020-transformers} library. We use linear schedule for learning rate decay. 
Maximum sequence length is set as 128 across tokenization, training, and fine-tuning.  Due to compute limitations, having higher maximum sequence length lead to out-of-memory errors.
Mini-batches are created by weighted sampling based on language priors with exponent $s=0.7$.
In LM pre-training, mini-batch of 48 samples and gradient accumulation of 53 is used making the effective batch size as 2,544\footnote{Loss curves of LM pre-training in supplementary material.}. Apex\footnote{\url{https://github.com/nvidia/apex}} library is used with \verb O1  optimization level to allow mixed precision training. 
In all our experiments on fine-tuning, we perform a grid-search with respect to \textit{learning-rate} and \textit{batch size} values of $\{ 1, 3, 5\}\times10^{-5}$ and $\{ 16, 32, 64\}$ respectively.

\begin{sidewaystable*}
\centering
.% For \textit{wnli-translated}, \textit{wikiann-ner}, and \textit{ud-pos} tasks we report F-Score. For other tasks we report Accuracy.\label{tab:mono-multi}}
\resizebox{\linewidth}{!}{%
\begin{tabular}{l|rrr|rrr|rrr|rrr|rrr} 
\toprule
 \textbf{LG}  & \multicolumn{3}{c|}{\textbf{mBERT} } & \multicolumn{3}{c|}{\textbf{IndicBERT} }  & \multicolumn{3}{c|}{\textbf{MuRIL} } &  \multicolumn{3}{c|}{\textbf{IA-Transliterated} } & \multicolumn{3}{c}{\textbf{IA-Original} } \\
 \cmidrule{2-16}
\textbf{FT $\rightarrow$}  & \textbf{Mono} & \textbf{Multi} & \multicolumn{1}{c|}{$\delta_{MB}$}& \textbf{Mono} & \textbf{Multi} & \multicolumn{1}{c|}{$\delta_{IB}$} &
\textbf{Mono} & \textbf{Multi} & \multicolumn{1}{c|}{$\delta_{MR}$}& \textbf{Mono} & \textbf{Multi} & \multicolumn{1}{c|}{$\delta_{TR}$} &
\textbf{Mono} & \textbf{Multi} &  \multicolumn{1}{c}{$\delta_O$} \\ 
\midrule
\multicolumn{16}{c}{\textbf{Textual Entailment / }copa-translated (Accuracy) } \\
hi & 0.6590 & 0.5909 & \textcolor{red}{(-10.33)}  & 0.6250 & \textbf{0.6705} & \textcolor{blue}{(+7.28)} & 0.5568 & 0.6477 & \textcolor{blue}{(+16.33)} & 0.6591 & 0.5455 & \textcolor{red}{(-17.24)} & 0.5796 & 0.5796 & (0.00)\\
gu & 0.4318 & 0.5795 & \textcolor{blue}{(+34.21)} & 0.5341 & \textbf{0.6591} & \textcolor{blue}{(+23.40)} & 0.5909 & 0.5795 & \textcolor{red}{(-1.93)} & 0.5909 & 0.6250  & \textcolor{blue}{(+5.77)} & 0.5113 & 0.5227 & \textcolor{blue}{(+2.23)} \\
mr & 0.5568 & 0.5909 & \textcolor{blue}{(+6.12)} & 0.5909 & \textbf{0.6591} & \textcolor{blue}{(+11.54)} & 0.5682 & 0.6363 & \textcolor{blue}{(+11.99)} & 0.5909 & 0.6023 & \textcolor{blue}{(+1.93)} & 0.6590 & 0.5454 & \textcolor{red}{(-17.24)} \\ 
\midrule
\multicolumn{16}{c}{\textbf{Textual Entailment / }wnli-translated (F-Score) } \\
hi & 0.3604 & 0.4020 & \textcolor{blue}{(+11.54)} & 0.3604 & \textbf{0.4855} & \textcolor{blue}{(+34.71)} & 0.3604 & 0.3604 & (0.00) & 0.3604 & 0.3604 & (0.00)& 0.3616 & 0.3604 & \textcolor{red}{(-0.33)} \\
gu & 0.3604 & 0.4123 & \textcolor{blue}{(+14.40)} & 0.3604 & \textbf{0.5071} & \textcolor{blue}{(+40.70)} & 0.3604 & 0.3604 & (0.00) & 0.3107 & 0.3604 & \textcolor{blue}{(+15.99)} & 0.3604 & 0.3604 & (0.00) \\
mr & 0.4167 & 0.4920 & \textcolor{blue}{(+18.07)} & 0.3604 & \textbf{0.4685} & \textcolor{blue}{(+29.99)} & 0.3914 & 0.3604 & \textcolor{red}{(-7.92)} & 0.4228 & 0.3576 & \textcolor{red}{(-15.42)} & 0.3604 & 0.3604 & (0.00) \\ 
\midrule
\multicolumn{16}{c}{\textbf{Entity Classification/ }wikiann-ner (F-Score) } \\
hi & 0.8656 & 0.9197 & \textcolor{blue}{(+6.25)} & 0.9031 & 0.9229 & \textcolor{blue}{(+2.19)} & 0.9237 & \textbf{0.9446} & \textcolor{blue}{(+2.26)} & 0.8720 & 0.8897 & \textcolor{blue}{(+2.03)} & 0.8476 & 0.8776 & \textcolor{blue}{(+3.54)} \\
bn & 0.9181 & 0.9671 & \textcolor{blue}{(+5.34)} & 0.9339 & \textbf{0.9654} & \textcolor{blue}{(+3.37)} & 0.9503 & 0.9645 & \textcolor{blue}{(+1.49)} & 0.9211 & 0.9397 & \textcolor{blue}{(+2.02)} & 0.9486 & 0.9557 & \textcolor{blue}{(+0.75)} \\
gu & 0.6804 & 0.8893 & \textcolor{blue}{(+30.70)} & 0.7021 & 0.9027 & \textcolor{blue}{(+28.57)} & 0.8016 & \textbf{0.9058} & \textcolor{blue}{(+12.99)} & 0.7306 & 0.7796 & \textcolor{blue}{(+6.71)} & 0.8318 & 0.8650 & \textcolor{blue}{(+3.99)} \\
mr & 0.9127 & 0.8882 & \textcolor{red}{(-2.68)} & 0.8871 & 0.8965 & \textcolor{blue}{(+1.06)} & 0.9199 & \textbf{0.9274} & \textcolor{blue}{(+0.82)} & 0.8675 & 0.8675 & (0.00) & 0.8482 & 0.8605 & \textcolor{blue}{(+1.45)} \\
or & 0.1905 & 0.3457 & \textcolor{blue}{(+81.47)} & 0.3509 & \textbf{0.8896} & \textcolor{blue}{(+153.52)} & 0.3882 & 0.8848 & \textcolor{blue}{(+127.92)} & 0.3460 & 0.6436 & \textcolor{blue}{(+86.01)} & 0.5737 & 0.7038 & \textcolor{blue}{(+22.68)} \\
pa & 0.5000 & 0.8667 & \textcolor{blue}{(+73.34)} & 0.4444 & 0.8593 & \textcolor{blue}{(+93.36)} & 0.8535 & \textbf{0.9086} & \textcolor{blue}{(+6.45)} & 0.3491 & 0.5789 & \textcolor{blue}{(+65.83)} & 0.6313 & 0.7456 & \textcolor{blue}{(+18.10)} \\ 
\midrule
\multicolumn{16}{c}{\textbf{Title Prediction/ }wiki-section-title (Accuracy) } \\
hi & 0.8012 & 0.8081 & \textcolor{blue}{(+0.86)} & 0.7780 & 0.7976 & \textcolor{blue}{(+2.52)} & 0.8528 & \textbf{0.8622} & \textcolor{blue}{(+1.10)} & 0.6779 & 0.6994 & \textcolor{blue}{(+3.17)} & 0.6761 & 0.6807 & \textcolor{blue}{(+0.68)} \\
bn & 0.8253 & 0.8285 & \textcolor{blue}{(+0.39)} & 0.8266 & 0.8157 & \textcolor{red}{(-1.32)} & 0.8781 & \textbf{0.8874} & \textcolor{blue}{(+1.06)} & 0.6135 & 0.7111 & \textcolor{blue}{(+15.91)} & 0.7062 & 0.7097 & \textcolor{blue}{(+0.49)} \\
gu & 0.7452 & 0.7858 & \textcolor{blue}{(+5.45)} & 0.6879 & 0.8074 & \textcolor{blue}{(+17.37)} & 0.8465 & \textbf{0.8745} & \textcolor{blue}{(+3.31)} & 0.2614 & 0.6970 & \textcolor{blue}{(+166.64)} & 0.4044 & 0.6890 & \textcolor{blue}{(+70.38)} \\
mr & 0.8049 & 0.8438 & \textcolor{blue}{(+4.83)} & 0.7744 & 0.8247 & \textcolor{blue}{(+6.49)} & 0.8529 & \textbf{0.8913} & \textcolor{blue}{(+4.50)} & 0.5299 & 0.7167 & \textcolor{blue}{(+35.25)} & 0.5031 & 0.7083 & \textcolor{blue}{(+40.78)} \\
or & 0.2222 & 0.2829 & \textcolor{blue}{(+27.31)} & 0.6825 & 0.8108 & \textcolor{blue}{(+18.79)} & 0.8167 & \textbf{0.8845} & \textcolor{blue}{(+8.30)} & 0.2928 & 0.7231 & \textcolor{blue}{(+146.96)} & 0.3147 & 0.6852 & \textcolor{blue}{(+117.73)} \\
pa & 0.7247 & 0.7648 & \textcolor{blue}{(+5.53)} & 0.7754 & 0.7903 & \textcolor{blue}{(+1.92)} & 0.8240 & \textbf{0.8487} & \textcolor{blue}{(+2.99)} & 0.2817 & 0.7064 & \textcolor{blue}{(+150.76)} & 0.6235 & 0.6800 & \textcolor{blue}{(+9.06)} \\ 
\midrule
\multicolumn{16}{c}{\textbf{Part-of-Speech Tagging / }ud-pos (F-Score) } \\
hi & 0.9693 & 0.9698 & \textcolor{blue}{(+0.05)} & 0.9755 & 0.9748 & \textcolor{red}{(-0.07)} & \textbf{0.9779} & 0.9778 & \textcolor{red}{(-0.01)} & 0.9618 & 0.9636 & \textcolor{blue}{(+0.19)} & 0.9562 & 0.9528 & \textcolor{red}{(-0.36)} \\
mr & \textbf{0.9024} & 0.8932 & \textcolor{red}{(-1.02)} & 0.8024 & 0.8886 & \textcolor{blue}{(+10.74)} & 0.5388 & 0.7478 & \textcolor{blue}{(+38.79)} & 0.8114 & 0.8290 & \textcolor{blue}{(+2.17)} & 0.7906 & 0.7926 & \textcolor{blue}{(+0.25)} \\
ur & 0.9102 & 0.8149 & \textcolor{red}{(-10.47)} & 0.9047 & 0.8249 & \textcolor{red}{(-8.82)} & \textbf{0.9168} & 0.8497 & \textcolor{red}{(-7.32)} & 0.9026 & 0.9080 & \textcolor{blue}{(+0.60)} & 0.8915 & 0.7858 & \textcolor{red}{(-11.85)} \\
\bottomrule
\end{tabular}
}
\caption{Comparison of multilingual fine-tuning vs monolingual fine-tuning of various Indo-Aryan LMs. The figure in parenthesis is relative difference $\delta$ (Eq. \ref{eq:delta})\label{tab:mono-multi}. The monolingual results for mBERT and IndicBERT models are as reported by \citet{kakwani2020inlpsuite} except for wnli-translated and POS-Tagging tasks. LG and FT stand for language and fine-tuning, respectively. }
\end{sidewaystable*}

\section{Experiments}

To answer the research questions, we experiment on a variety of tasks suitable for multilingual fine-tuning and analyse the results. 
To investigate RQ1, in \S\ref{sec:multi}, the first set of experiments are aimed to understand the utility of multilingual FT with related languages.
To investigate RQ2, in \S\ref{sec:grad}, the second set of experiments are designed to track gradual performance variation with addition of assisting languages.
With last set of analysis, in \S\ref{sec:tr}, we investigate the role of transliteration.

\subsection{Effectiveness on Multilingual Tasks\label{sec:multi}}
We experiment on four tasks suitable for multilingual fine-tuning protocol, including three from IndicGLUE \cite{kakwani-etal-2020-indicnlpsuite} and POS tagging \cite{11234/1-3424}.
\begin{enumerate}

    % \noindent
    \item \textbf{Textual Entailment} task on copa-translated and wnli-translated dataset (Hi, Gu, Mr)

    % \noindent
    \item \textbf{Title Prediction} task on wiki-section-title dataset (Hi, Bn, Gu, Mr, Or, Pa)

    % \noindent
    \item \textbf{Named Entity Recognition} task on wikiann-ner dataset (Hi, Bn, Gu, Mr, Or, Pa)
    % \item Article-Genre Classification task on inltk-headline dataset (on Gujarati and Marathi)
    % \item 
    
    % \noindent
    \item \textbf{Part-of-Speech Tagging} task on Universal Dependency datasets (Hi, Mr, Ur)
\end{enumerate}
 
\noindent We do not show results on Cloze-style Question Answering task of IndicGLUE as it is meant to evaluate masked-token prediction of an LM, and does not involve downstream task training.

 We utilize mBERT, IndicBERT, MuRIL, IndoAryan-Original (IA-O) and IndoAryan-Transliterated (IA-TR) -- last two being pre-trained by us as detailed in \S\ref{sec:lm}.
 All the five LMs are fine-tuned in monolingual and multilingual modes, to pursue investigation for RQ 1.
 Only the IA-TR model is fine-tuned with transliterated versions of the downstream task data; remaining four models are fine-tuned with original script downstream task data.
  The results of this set of experiments are reported in Table \ref{tab:mono-multi}. Along with absolute metrics, the relative difference between mono- and multi- lingual fine-tuning (FT) is also reported. The relative difference is calculated as 
 \begin{equation}
 \delta = 100\times\frac{M_{multi}-M_{mono}}{M_{mono}}  \label{eq:delta}
 \end{equation}
 where $M_{mono}$ and $M_{multi}$ are performance measures of monolingual and multilingual fine-tuning respectively. Key observations are as following:

\begin{figure}[!t]
\footnotesize
    \centering
    
   \includegraphics[width=0.45\linewidth]{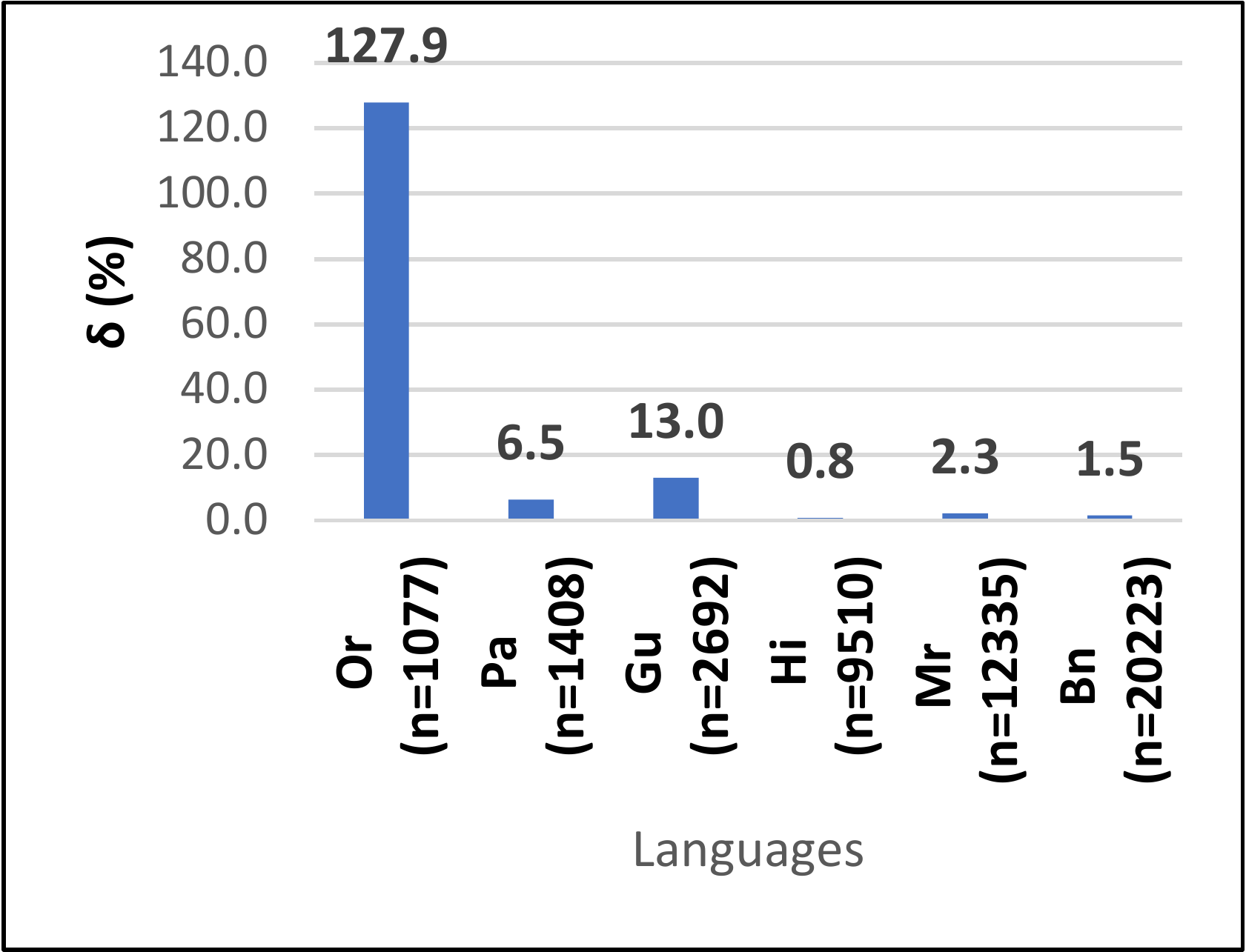}
    \includegraphics[width=0.45\linewidth]{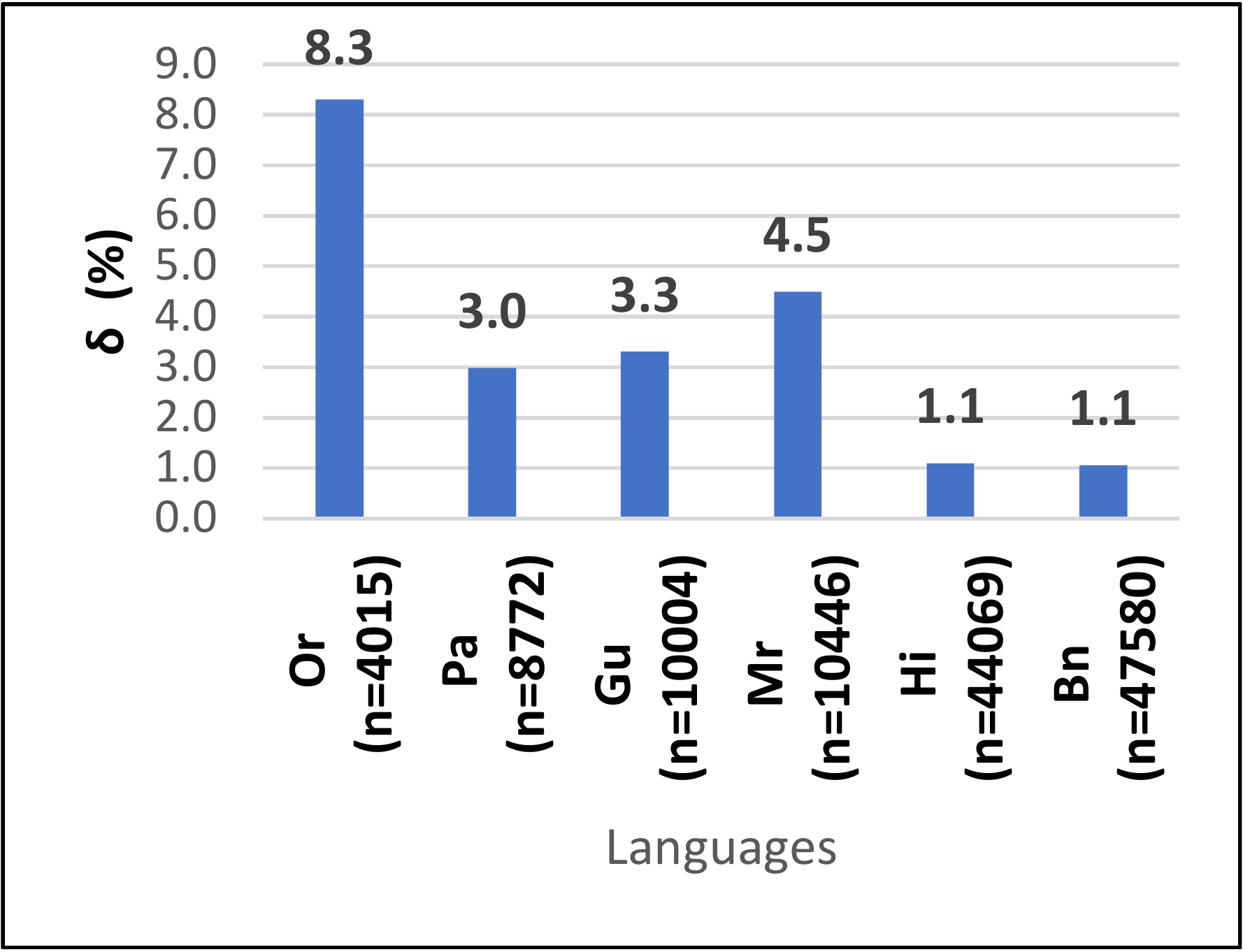}
    
    \caption{Relative performance difference $\delta$ (Eq. \ref{eq:delta}) for MuRIL model on (left) NER and (right) title prediction tasks. $n$ is number of training samples}. 
    
    \label{fig:low-resource}
\end{figure}
\ \\
   
  \noindent \textbf{Monolingual vs Multilingual Fine-Tuning:}
 In this analysis, higher the $\delta$, the stronger is the answer to RQ1 in affirmation. In the Table \ref{tab:mono-multi} the positive $\delta$, shown in blue color, indicates the cases where multilingual FT improves over monolingual FT.
 It can be observed that for languages with limited labelled data for the downstream task, the multilingual fine-tuning is resulting in enormous improvements. Across all the five LMs, the trend is consistent.
 For example, on wikiann-ner task, F-Score on Oriya  improved from 0.3882 and 0.3460 to 0.8848 and 0.6436, respectively for MuRIL and IA-TR models; while significant improvements are seen in other languages too. Similar trend is visible in wiki-section-title prediction task too, where improvements are seen for all the languages.
 Broadly, across LMs, tasks, and languages, the multilingual FT shows improvement over monolingual FT.
 This helps formulate the answer to RQ1 as \textit{the multilingual fine-tuning with related languages can yield huge (up to 40\% on absolute scale) improvement for low-resource languages (such as Oriya and Punjabi), and statistically significant (up to 10\%) improvements on high resource languages (such as Hindi and Bengali), depending on the task.} 
 
\begin{figure}[t]
     \centering
     \includegraphics[height=1.7cm]{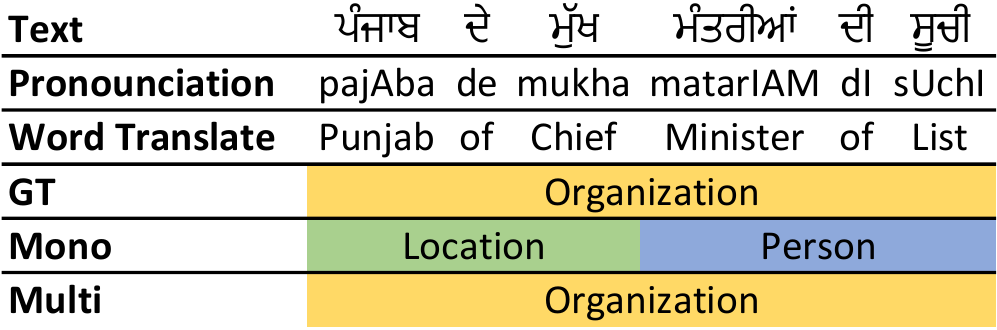}\\
     (a) Entity tag change\\
     \includegraphics[height=1.7cm]{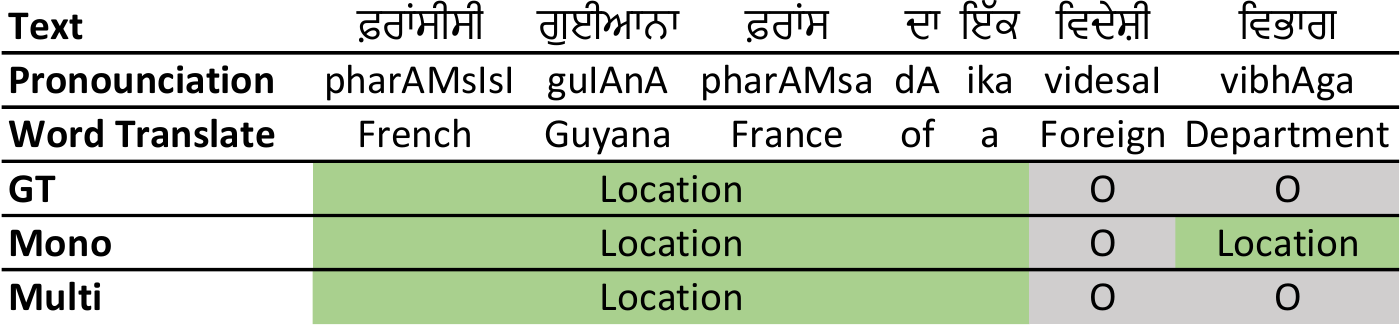}\\
     (b) Spurious entity extraction rectified \\
     \caption{Examples of prediction improvement by multilingual FT compared to monolingual FT, for NER task for two examples of Punjabi written in Gurmukhi script. }
     \label{fig:eg}
 \end{figure}

 Note, that this is in contrast to the observations of \citet{tsai-etal-2019-small} and \citet{kondratyuk-2019-cross} that indicate slightly poor performance with multilingual fine-tuning. They fine-tune with more than forty languages together, without considering language relatedness. We observe large improvements by selecting only the languages of the family for multilingual fine-tuning. 
%  \textbf{todo: wnli numbers poor -- dataset limitation, bias}

 \ \\
 \noindent \textbf{Trade-off or Win-Win?:}
 Figure \ref{fig:low-resource} visualizes the the improvements by multilingual fine-tuning relative to the monolingual one along with the task training set size. It is clearly evident, the smaller the task training data set, the higher is the relative improvement. Arguably, the data limitation of a low resource language is abridged by the related high resource languages. These improvements are not at the cost of trading-off on high resource languages; it is win-win for all languages. In fact, a decrease in performance ($\delta <0$), indicated with red color, is observed in only 16 out of the total 105 (21 task-language pairs $\times$ 5 LMs) comparisons. 
 Interestingly, there is no task-language pair in which $\delta$ values corresponding to all the five LMs are negative, i.e. for all task-language pairs at least one LM always showed improvement using multilingual fine-tuning. Figure \ref{fig:eg} illustrates the types of improvements in predicting entity tags.

 \ \\
 \noindent \textbf{Best LM Across Tasks?:}
 Arguably, it is unfair to compare the pre-trained LMs due to vast difference in the number of languages they are pre-trained for (in range from 11 to 104), the size of the corpora, nature of the corpora (only monolingual vs parallel corpora), model types (RoBERTA, AlBERT, BERT), number of layers (8 or 12), tokenization, pre-training objectives, and compute consumed in training. Further, mBERT model is not pre-trained with Oriya. However, it is natural to inquire if there is a clear winner LM in the experimentation. The boldface figures in Table \ref{tab:mono-multi} shows the best results per task per language. 
 Most of the best metrics fall under either IndicBERT (for Textual Entailment tasks) or MuRIL models (for Title Prediction and mostly for Entity Classification tasks). The lowest performance obtained is with IA-O and IA-TR. 

 \begin{table*}[!t]
    % \footnotesize
     \centering
    %  \subfloat[\label{tab:ablation}
    %  \vspace{4pt}
    %  \subfloat[\label{tab:ablation}]
    \resizebox{0.65\textwidth}{!}
    {
    \large
        \begin{tabular}{lrc}
        \toprule
            \textbf{Train Set} & \textbf{Size} &
            \multicolumn{1}{c}{ \textbf{F-Score} }   \\ 
    \midrule
            % Train Set & Size &F1 on Oriya\\\midrule
          or & 1078 & 0.3882   \\\midrule
          \multicolumn{3}{l}{Base set: or} \\
          \cmidrule{1-2}
          \ \bf + gu  & 3425 & 0.9245 \\
          \ + bn  & 21379 & 0.8836  \\
          \ + hi  & 10590 & 0.8795 \\
          \ + pa  & 2487  & 0.8657  \\
          \ + mr  & 13415 & 0.8649  \\
          \midrule
          \multicolumn{3}{l}{Base set: or+gu} \\
          \cmidrule{1-2}
          \ \bf + bn & 23725 &  0.9301   \\
          \ + pa & 4884 & 0.9231   \\
          \ + hi & 12936 & 0.8836 \\
          \ + mr & 15761 & 0.8855  \\ \midrule
          \multicolumn{3}{l}{Base set: or+gu+bn} \\
          \cmidrule{1-2}
          \ \bf + mr & 36061 & 0.9151 \\
          \ + pa & 25184 & 0.9036\\
          \ + hi & 33236 & 0.8916  \\ \midrule
          \multicolumn{3}{l}{Base Set: or+gu+bn+mr} \\
          \cmidrule{1-2}
          \ \bf + hi & 45572 & \bf 0.9419  \\
          \ + pa & 37520 & 0.8922  \\
           \midrule
        %   \multicolumn{3}{l}{Base Set: or+gu+bn+mr+hi} \\
        %   \cmidrule{1-2}
        %   \ + en & 55572 & 0.9042  \\
        %   \midrule
          All & 46665 & 0.8848  \\\bottomrule
          
     \end{tabular}%
    \quad
    %  }
    %  \resizebox{0.4\textwidth}{!}
    %  {
     \begin{tabular}{lrc}
     \toprule
            \textbf{Train Set} & \textbf{Size} &
            \multicolumn{1}{c}{ \textbf{F-Score} }   \\ 
    \midrule
          pa  & 1409 & 0.8535  \\ \midrule
          \multicolumn{3}{l}{Base set: pa} \\
          \cmidrule{1-2}
          \ \bf + bn  & 21579 & 0.9156  \\
          \ + mr  & 13795 &  0.8883    \\
          \ + hi  & 10970 & 0.8759   \\
          \ + gu   & 3805 &  0.8673   \\
          \ + or  & 2487 & 0.8426   \\
          \midrule
          \multicolumn{3}{l}{Base set: pa+bn} \\
          \cmidrule{1-2}
          \ \bf + hi & 31270 & 0.9286   \\
          \ + mr & 34095 & 0.9160  \\
          \ + or & 22838 & 0.9137    \\
          \ + gu & 24105 & 0.9105   \\\midrule
          \multicolumn{3}{l}{Base set: pa+bn+hi} \\
          \cmidrule{1-2}
          \ \bf + mr & 43606 & 0.9211   \\
          \ + or & 32349 & 0.9156   \\
          \ + gu & 33616 & 0.9132  \\
          \midrule
          \multicolumn{3}{l}{Base set: pa+bn+hi+mr}\\
          \cmidrule{1-2}
          \ \bf + gu & 45952 & \bf 0.9567  \\
          \ + or & 44685 & 0.9231   \\
          \midrule
        %   \multicolumn{3}{l}{Base Set: pa+bn+hi+mr+gu} \\
        %   \cmidrule{1-2}
        %   \ + en & 54685 &   \\
        %   \midrule
          All & 46665 &  0.9086 \\ \bottomrule
     \end{tabular}%
     }
     \caption{Study  of graded addition of languages for NER task on low resource languages of (left) Oriya and (right) Punjabi using MuRIL. The scheme of adding the languages is similar to Greedy Forward Selection of features in Machine Learning.  \label{tab:ablation} }
 \end{table*}
 
 \subsection{Gradation of Multilinguality\label{sec:grad}}
 We further dwell into understanding whether the degree of improvement varies by the language closeness within language family. 
 Specifically, we start with the monolingual training, i.e. training set contains only the target language. 
 Then, we experiment by adding each related language to the training set separately.
 The language that yields highest performance boost is selected for adding to the training set. Thus, a new training set consisting of two languages is obtained. This is repeated until all the related languages are added to the training set, resulting in the all-language multilingual FT.
 This approach is similar to \emph{Sequential Forward Selection of Features} in machine learning.
 Further, we relate the subfamily categorization of IA family for this analysis.

 Experiments are performed for NER task on MuRIL model, with Oriya and Punjabi as the evaluation language. The results are reported in Table \ref{tab:ablation}. 
 For detailed discussion, consider the case of Oriya, on one end of the spectrum, in the first row, we have a monolingual fine-tuned model with only Oriya, whereas on the other end of the spectrum, in the last row, we have a multilingually fine-tuned model with Oriya, Bengali, Hindi, Gujarati, Marathi, and Punjabi. In the middle span, 
 we have Oriya aided by each -- Bengali, Punjabi, Marathi, Gujarati, and Hindi, separately; and their combinations.
 Since adding Gujarati to Oriya yields best result compared to adding any other language, \textit{or+gu} is taken as the base training set for next iteration. In the next iteration, adding Bengali to \textit{or+gu} provides highest boost, thus \textit{or+gu+bn} forms the base iteration for next iteration. A similar exercise is performed with Punjabi as the base language too.
 
  In case of Oriya, adding the Gujarati data resulted in about 54 percentage points improvement (38.8\% to 92.4\%), which is further improved  by about 0.5\% with addition of Bengali (93.0\%). It appears that Hindi, Punjabi, and Marathi each negatively interferes with \textit{or+gu+bn} set, resulting in at least 1.5 percentage points performance drop. The best performance of 94.19\% is obtained with the \textit{or+gu+bn+mr+hi} set (i.e. all but Punjabi), which is 5.7 percentage points higher than considering all the languages together. 
  
  It is natural to ask if Punjabi has negative interference with Oriya for the task. \textit{or+pa} yields 47.7 percentage points improvement over only Oriya, which indicates positive interference between them. Further, \textit{or+gu} and \textit{or+gu+pa} are almost similar (0.9245 and 0.9231) indicates that perhaps, \textit{pa} is redundant to \textit{gu} for assisting \textit{or} on the task. However, adding \textit{pa} to \textit{or+gu+bn}, \textit{or+gu+bn+mr}, and \textit{or+gu+bn+mr+hi} results in 2-5\% drop; the common denominator being Bengali, it seems that Punjabi harms the most when the base set contains Bengali. Arguably, it indicates negative interference between Bengali and Punjabi for the task. 
  
 Also, note the improvements are not correlated with increase in the training set size; instead the smaller sets (e.g. \textit{or+gu} with 3425 samples) yields better results than larger sets (e.g. \textit{or+bn} of 21379).
 Therefore, gradual deviations should be credited to the addition language rather than the training set inflation. 
 overall, the answer to RQ 2 emerges \emph{within the set of related languages, likely, there exists a subset of languages that yields the best performance.}

 \subsection{Transliteration\label{sec:tr}}
 Next, we present a set of observations pertaining to the utility of transliteration to leverage the script similarity between the Indo-Aryan languages. 
 For a fair comparison, the IA-Original and IA-Transliterated models are considered as both of them are pre-trained by us on the original script and transliterated script versions of the same corpora.
 Thus, in this part of analysis, higher the $\delta_{TR}-\delta_{O}$ in Table \ref{tab:mono-multi}, the stronger is the role of explicit script normalization.
 
 \ \\
 \noindent\textbf{Transliteration with Multilingual FT:}
 Comparing the relative difference ($\delta$) for transliterated and original script models, it is observed that in 16 out of 21 task-language pairs $\delta_{TR} > \delta_{O}$;
 noteworthy is $\delta_{TR}=146.96\%$, and $\delta_{O}=117.73\%$ for Oriya language on wiki-section-title prediction task.
 It suggests that multilingual FT is even more suitable with transliteration. 
  Based on these results, the role of common script representation emerges as \textit{effectiveness of multilingual fine-tuning is significantly enhanced when coupled with common-script representation via transliteration.} 
  
\ \\
\noindent \textbf{Transliteration for LM:}
  Comparing performance of monolingual fine-tuning of the original script LM (IA-O) and the transliterated script LM (IA-TR) reveals that the later is better in only few (8 out of 21) experiments.
 This is somewhat counter intuitive, as the common script representation should have made the LM pre-train better due to the presence of cognates. We speculate following two rationales.
 \begin{itemize}
     \item  Firstly, it indicates that, perhaps, even without explicit alignment of cognates (via transliteration) the model is able to align their embeddings implicitly, corroborating with \cite{conneau-etal-2020-emerging,pires-etal-2019-multilingual}.
    \item Secondly, the byte-Level BPE and the unicode block arrangements for Indo-Aryan languages may be at play underneath this phenomena also. For example, the consonant Pa in Hindi \indicWords{devaPa} (0xe0 0xa4 0xaa), Oriya \indicWords{oriaPa} (0xe0 0xac  0xaa), and Punjabi \indicWords{punPa} (0xe0 0xa8 0xaa) are apart by their unicode block offset differences. Thus, potentially, a model knowing the byte level representations of the writing system could learn to map them, provided the loss function guides it.
 \end{itemize}
 However, we leave the further inquiry into the exact phenomena for the future work.

\section{Conclusion}
We show that multilingual fine-tuning efficiently leverages language relatedness leading to improvements over monolingual approach. We substantiate this claim on the Indo-Aryan language family with experiments on five language models. Multilingual fine-tuning is particularly effective for low-resource languages (e.g., Oriya and Punjabi show improvement up to 150\% on relative scale). 
Also, we show that careful selection of subset of related languages, can further improve performance. 
Devising automatic approaches for finding optimal subset of related languages is a promising future direction.
Additionally, in multilingual fine-tuning, we see some benefits of transliteration to common script. 

\bibliography{anthology,acl2020}

\begin{thebibliography}{36}
\expandafter\ifx\csname natexlab\endcsname\relax\def\natexlab#1{#1}\fi

\bibitem[{Bhat et~al.(2015)Bhat, Mujadia, Tammewar, Bhat, and
  Shrivastava}]{Bhat:2014:ISS:2824864.2824872}
Irshad~Ahmad Bhat, Vandan Mujadia, Aniruddha Tammewar, Riyaz~Ahmad Bhat, and
  Manish Shrivastava. 2015.
\newblock \href {https://doi.org/10.1145/2824864.2824872} {{IIIT-H System
  Submission for FIRE2014 Shared Task on Transliterated Search}}.
\newblock In \emph{Proceedings of the Forum for Information Retrieval
  Evaluation}, FIRE '14, pages 48--53, New York, NY, USA. ACM.

\bibitem[{Bhattacharyya et~al.(2016)Bhattacharyya, Khapra, and
  Kunchukuttan}]{bhattacharyya-etal-2016-statistical}
Pushpak Bhattacharyya, Mitesh~M. Khapra, and Anoop Kunchukuttan. 2016.
\newblock \href {https://doi.org/10.18653/v1/N16-4006} {Statistical machine
  translation between related languages}.
\newblock In \emph{Proceedings of the 2016 Conference of the North {A}merican
  Chapter of the Association for Computational Linguistics: Tutorial
  Abstracts}, pages 17--20, San Diego, California. Association for
  Computational Linguistics.

\bibitem[{Bojar et~al.(2014)Bojar, Diatka, Stra{\v n}{\'a}k, Tamchyna, and
  Zeman}]{11858/00-097C-0000-0023-625F-0}
Ond{\v r}ej Bojar, Vojt{\v e}ch Diatka, Pavel Stra{\v n}{\'a}k, Ale{\v s}
  Tamchyna, and Daniel Zeman. 2014.
\newblock \href {http://hdl.handle.net/11858/00-097C-0000-0023-625F-0}
  {{HindEnCorp} 0.5}.
\newblock {LINDAT}/{CLARIAH}-{CZ} digital library at the Institute of Formal
  and Applied Linguistics ({{\'U}FAL}), Faculty of Mathematics and Physics,
  Charles University.

\bibitem[{Brown et~al.(2020)Brown, Mann, Ryder, Subbiah, Kaplan, Dhariwal,
  Neelakantan, Shyam, Sastry, Askell, Agarwal, Herbert-Voss, Krueger, Henighan,
  Child, Ramesh, Ziegler, Wu, Winter, Hesse, Chen, Sigler, Litwin, Gray, Chess,
  Clark, Berner, McCandlish, Radford, Sutskever, and
  Amodei}]{brown2020language}
Tom~B. Brown, Benjamin Mann, Nick Ryder, Melanie Subbiah, Jared Kaplan,
  Prafulla Dhariwal, Arvind Neelakantan, Pranav Shyam, Girish Sastry, Amanda
  Askell, Sandhini Agarwal, Ariel Herbert-Voss, Gretchen Krueger, Tom Henighan,
  Rewon Child, Aditya Ramesh, Daniel~M. Ziegler, Jeffrey Wu, Clemens Winter,
  Christopher Hesse, Mark Chen, Eric Sigler, Mateusz Litwin, Scott Gray,
  Benjamin Chess, Jack Clark, Christopher Berner, Sam McCandlish, Alec Radford,
  Ilya Sutskever, and Dario Amodei. 2020.
\newblock \href {http://arxiv.org/abs/2005.14165} {Language models are few-shot
  learners}.

\bibitem[{Conneau et~al.(2020{\natexlab{a}})Conneau, Khandelwal, Goyal,
  Chaudhary, Wenzek, Guzm{\'a}n, Grave, Ott, Zettlemoyer, and
  Stoyanov}]{conneau-etal-2020-unsupervised}
Alexis Conneau, Kartikay Khandelwal, Naman Goyal, Vishrav Chaudhary, Guillaume
  Wenzek, Francisco Guzm{\'a}n, Edouard Grave, Myle Ott, Luke Zettlemoyer, and
  Veselin Stoyanov. 2020{\natexlab{a}}.
\newblock \href {https://doi.org/10.18653/v1/2020.acl-main.747} {Unsupervised
  cross-lingual representation learning at scale}.
\newblock In \emph{Proceedings of the 58th Annual Meeting of the Association
  for Computational Linguistics}, pages 8440--8451, Online. Association for
  Computational Linguistics.

\bibitem[{Conneau et~al.(2020{\natexlab{b}})Conneau, Wu, Li, Zettlemoyer, and
  Stoyanov}]{conneau-etal-2020-emerging}
Alexis Conneau, Shijie Wu, Haoran Li, Luke Zettlemoyer, and Veselin Stoyanov.
  2020{\natexlab{b}}.
\newblock \href {https://doi.org/10.18653/v1/2020.acl-main.536} {Emerging
  cross-lingual structure in pretrained language models}.
\newblock In \emph{Proceedings of the 58th Annual Meeting of the Association
  for Computational Linguistics}, pages 6022--6034, Online. Association for
  Computational Linguistics.

\bibitem[{Devlin et~al.(2019)Devlin, Chang, Lee, and
  Toutanova}]{devlin2019bert}
Jacob Devlin, Ming-Wei Chang, Kenton Lee, and Kristina Toutanova. 2019.
\newblock Bert: Pre-training of deep bidirectional transformers for language
  understanding.
\newblock In \emph{Proceedings of the 2019 Conference of the North American
  Chapter of the Association for Computational Linguistics: Human Language
  Technologies, Volume 1 (Long and Short Papers)}, pages 4171--4186.

\bibitem[{Dolicki and Spanakis(2021)}]{dolicki2021analysing}
Błażej Dolicki and Gerasimos Spanakis. 2021.
\newblock \href {http://arxiv.org/abs/2105.05975} {{Analysing The Impact Of
  Linguistic Features On Cross-Lingual Transfer}}.

\bibitem[{Jain et~al.(2020)Jain, Deshpande, Shridhar, Laumann, and
  Dash}]{jain2020indic}
Kushal Jain, Adwait Deshpande, Kumar Shridhar, Felix Laumann, and Ayushman
  Dash. 2020.
\newblock {Indic-Transformers: An Analysis of Transformer Language Models for
  Indian Languages}.
\newblock \emph{arXiv preprint arXiv:2011.02323}.

\bibitem[{Jawaid et~al.(2014)Jawaid, Kamran, and
  Bojar}]{11858/00-097C-0000-0023-65A9-5}
Bushra Jawaid, Amir Kamran, and Ond{\v r}ej Bojar. 2014.
\newblock \href {http://hdl.handle.net/11858/00-097C-0000-0023-65A9-5} {Urdu
  monolingual corpus}.
\newblock {LINDAT}/{CLARIAH}-{CZ} digital library at the Institute of Formal
  and Applied Linguistics ({{\'U}FAL}), Faculty of Mathematics and Physics,
  Charles University.

\bibitem[{Kakwani et~al.(2020{\natexlab{a}})Kakwani, Kunchukuttan, Golla,
  Gokul, Bhattacharyya, Khapra, and Kumar}]{kakwani2020inlpsuite}
Divyanshu Kakwani, Anoop Kunchukuttan, Satish Golla, NC~Gokul, Avik
  Bhattacharyya, Mitesh~M Khapra, and Pratyush Kumar. 2020{\natexlab{a}}.
\newblock inlpsuite: Monolingual corpora, evaluation benchmarks and pre-trained
  multilingual language models for indian languages.
\newblock In \emph{Proceedings of the 2020 Conference on Empirical Methods in
  Natural Language Processing: Findings}, pages 4948--4961.

\bibitem[{Kakwani et~al.(2020{\natexlab{b}})Kakwani, Kunchukuttan, Golla, N.C.,
  Bhattacharyya, Khapra, and Kumar}]{kakwani-etal-2020-indicnlpsuite}
Divyanshu Kakwani, Anoop Kunchukuttan, Satish Golla, Gokul N.C., Avik
  Bhattacharyya, Mitesh~M. Khapra, and Pratyush Kumar. 2020{\natexlab{b}}.
\newblock \href {https://www.aclweb.org/anthology/2020.findings-emnlp.445}
  {{I}ndic{NLPS}uite: Monolingual corpora, evaluation benchmarks and
  pre-trained multilingual language models for {I}ndian languages}.
\newblock In \emph{Findings of the Association for Computational Linguistics:
  EMNLP 2020}, pages 4948--4961, Online. Association for Computational
  Linguistics.

\bibitem[{Khanuja et~al.(2021)Khanuja, Bansal, Mehtani, Khosla, Dey, Gopalan,
  Margam, Aggarwal, Nagipogu, Dave, Gupta, Gali, Subramanian, and
  Talukdar}]{khanuja2021muril}
Simran Khanuja, Diksha Bansal, Sarvesh Mehtani, Savya Khosla, Atreyee Dey,
  Balaji Gopalan, Dilip~Kumar Margam, Pooja Aggarwal, Rajiv~Teja Nagipogu,
  Shachi Dave, Shruti Gupta, Subhash Chandra~Bose Gali, Vish Subramanian, and
  Partha Talukdar. 2021.
\newblock \href {http://arxiv.org/abs/2103.10730} {{MuRIL: Multilingual
  Representations for Indian Languages}}.

\bibitem[{Kondratyuk(2019)}]{kondratyuk-2019-cross}
Dan Kondratyuk. 2019.
\newblock \href {https://doi.org/10.18653/v1/W19-4203} {Cross-lingual
  lemmatization and morphology tagging with two-stage multilingual {BERT}
  fine-tuning}.
\newblock In \emph{Proceedings of the 16th Workshop on Computational Research
  in Phonetics, Phonology, and Morphology}, pages 12--18, Florence, Italy.
  Association for Computational Linguistics.

\bibitem[{Kulshreshtha et~al.(2020)Kulshreshtha, Garcia, and
  Chang}]{kulshreshtha2020cross}
Saurabh Kulshreshtha, Jose Luis~Redondo Garcia, and Ching~Yun Chang. 2020.
\newblock {Cross-lingual Alignment Methods for Multilingual BERT: A Comparative
  Study}.
\newblock In \emph{Proceedings of the 2020 Conference on Empirical Methods in
  Natural Language Processing: Findings}, pages 933--942.

\bibitem[{Kumar et~al.(2018)Kumar, Lahiri, Alok, Ojha, Jain, Basit, and
  Dawer}]{kumar2018automatic}
Ritesh Kumar, Bornini Lahiri, Deepak Alok, Atul~Kr Ojha, Mayank Jain, Abdul
  Basit, and Yogesh Dawer. 2018.
\newblock {Automatic identification of closely-related Indian languages:
  Resources and experiments}.
\newblock \emph{arXiv preprint arXiv:1803.09405}.

\bibitem[{Kumar et~al.(2020)Kumar, Kumar, Kanojia, and
  Bhattacharyya}]{kumar-etal-2020-passage}
Saurav Kumar, Saunack Kumar, Diptesh Kanojia, and Pushpak Bhattacharyya. 2020.
\newblock \href {https://www.aclweb.org/anthology/2020.sltu-1.49} {{``}a
  passage to {I}ndia{''}: Pre-trained word embeddings for {I}ndian languages}.
\newblock In \emph{Proceedings of the 1st Joint Workshop on Spoken Language
  Technologies for Under-resourced languages (SLTU) and Collaboration and
  Computing for Under-Resourced Languages (CCURL)}, pages 352--357, Marseille,
  France. European Language Resources association.

\bibitem[{Kunchukuttan et~al.(2018)Kunchukuttan, Mehta, and
  Bhattacharyya}]{kunchukuttan-etal-2018-iit}
Anoop Kunchukuttan, Pratik Mehta, and Pushpak Bhattacharyya. 2018.
\newblock \href {https://www.aclweb.org/anthology/L18-1548} {The {IIT} {B}ombay
  {E}nglish-{H}indi parallel corpus}.
\newblock In \emph{Proceedings of the Eleventh International Conference on
  Language Resources and Evaluation ({LREC}-2018)}, Miyazaki, Japan. European
  Languages Resources Association (ELRA).

\bibitem[{Kunchukuttan et~al.(2015)Kunchukuttan, Puduppully, and
  Bhattacharyya}]{kunchukuttan-etal-2015-brahmi}
Anoop Kunchukuttan, Ratish Puduppully, and Pushpak Bhattacharyya. 2015.
\newblock \href {https://doi.org/10.3115/v1/N15-3017} {Brahmi-net: A
  transliteration and script conversion system for languages of the {I}ndian
  subcontinent}.
\newblock In \emph{Proceedings of the 2015 Conference of the North {A}merican
  Chapter of the Association for Computational Linguistics: Demonstrations},
  pages 81--85, Denver, Colorado. Association for Computational Linguistics.

\bibitem[{Lample and Conneau(2019)}]{lample2019cross}
Guillaume Lample and Alexis Conneau. 2019.
\newblock Cross-lingual language model pretraining.
\newblock \emph{arXiv preprint arXiv:1901.07291}.

\bibitem[{Lewis et~al.(2020)Lewis, Liu, Goyal, Ghazvininejad, Mohamed, Levy,
  Stoyanov, and Zettlemoyer}]{lewis-etal-2020-bart}
Mike Lewis, Yinhan Liu, Naman Goyal, Marjan Ghazvininejad, Abdelrahman Mohamed,
  Omer Levy, Veselin Stoyanov, and Luke Zettlemoyer. 2020.
\newblock \href {https://doi.org/10.18653/v1/2020.acl-main.703} {{BART}:
  Denoising sequence-to-sequence pre-training for natural language generation,
  translation, and comprehension}.
\newblock In \emph{Proceedings of the 58th Annual Meeting of the Association
  for Computational Linguistics}, pages 7871--7880, Online. Association for
  Computational Linguistics.

\bibitem[{Liu et~al.(2020)Liu, Gu, Goyal, Li, Edunov, Ghazvininejad, Lewis, and
  Zettlemoyer}]{liu2020multilingual}
Yinhan Liu, Jiatao Gu, Naman Goyal, Xian Li, Sergey Edunov, Marjan
  Ghazvininejad, Mike Lewis, and Luke Zettlemoyer. 2020.
\newblock Multilingual denoising pre-training for neural machine translation.
\newblock \emph{Transactions of the Association for Computational Linguistics},
  8:726--742.

\bibitem[{Liu et~al.(2019)Liu, Ott, Goyal, Du, Joshi, Chen, Levy, Lewis,
  Zettlemoyer, and Stoyanov}]{liu2019roberta}
Yinhan Liu, Myle Ott, Naman Goyal, Jingfei Du, Mandar Joshi, Danqi Chen, Omer
  Levy, Mike Lewis, Luke Zettlemoyer, and Veselin Stoyanov. 2019.
\newblock Roberta: A robustly optimized bert pretraining approach.
\newblock \emph{arXiv preprint arXiv:1907.11692}.

\bibitem[{Pires et~al.(2019)Pires, Schlinger, and
  Garrette}]{pires-etal-2019-multilingual}
Telmo Pires, Eva Schlinger, and Dan Garrette. 2019.
\newblock \href {https://doi.org/10.18653/v1/P19-1493} {How multilingual is
  multilingual {BERT}?}
\newblock In \emph{Proceedings of the 57th Annual Meeting of the Association
  for Computational Linguistics}, pages 4996--5001, Florence, Italy.
  Association for Computational Linguistics.

\bibitem[{Radford et~al.(2018)Radford, Narasimhan, Salimans, and
  Sutskever}]{radford2018improving}
Alec Radford, Karthik Narasimhan, Tim Salimans, and Ilya Sutskever. 2018.
\newblock Improving language understanding by generative pre-training (2018).
\newblock \emph{URL https://s3-us-west-2. amazonaws.
  com/openai-assets/research-covers/language-unsupervised/language\_
  understanding\_paper. pdf}.

\bibitem[{Radford et~al.(2019)Radford, Wu, Child, Luan, Amodei, and
  Sutskever}]{radford2019language}
Alec Radford, Jeff Wu, Rewon Child, David Luan, Dario Amodei, and Ilya
  Sutskever. 2019.
\newblock Language models are unsupervised multitask learners.

\bibitem[{Tran and Bisazza(2019)}]{tran-bisazza-2019-zero}
Ke~Tran and Arianna Bisazza. 2019.
\newblock \href {https://doi.org/10.18653/v1/D19-6132} {Zero-shot dependency
  parsing with pre-trained multilingual sentence representations}.
\newblock In \emph{Proceedings of the 2nd Workshop on Deep Learning Approaches
  for Low-Resource NLP (DeepLo 2019)}, pages 281--288, Hong Kong, China.
  Association for Computational Linguistics.

\bibitem[{Tsai et~al.(2019)Tsai, Riesa, Johnson, Arivazhagan, Li, and
  Archer}]{tsai-etal-2019-small}
Henry Tsai, Jason Riesa, Melvin Johnson, Naveen Arivazhagan, Xin Li, and Amelia
  Archer. 2019.
\newblock \href {https://doi.org/10.18653/v1/D19-1374} {Small and practical
  {BERT} models for sequence labeling}.
\newblock In \emph{Proceedings of the 2019 Conference on Empirical Methods in
  Natural Language Processing and the 9th International Joint Conference on
  Natural Language Processing (EMNLP-IJCNLP)}, pages 3632--3636, Hong Kong,
  China. Association for Computational Linguistics.

\bibitem[{Vaswani et~al.(2017)Vaswani, Shazeer, Parmar, Uszkoreit, Jones,
  Gomez, Kaiser, and Polosukhin}]{vaswani2017attention}
Ashish Vaswani, Noam Shazeer, Niki Parmar, Jakob Uszkoreit, Llion Jones,
  Aidan~N Gomez, {\L}ukasz Kaiser, and Illia Polosukhin. 2017.
\newblock Attention is all you need.
\newblock In \emph{Advances in neural information processing systems}, pages
  5998--6008.

\bibitem[{Wang et~al.(2020{\natexlab{a}})Wang, Cho, and Gu}]{wang2020neural}
Changhan Wang, Kyunghyun Cho, and Jiatao Gu. 2020{\natexlab{a}}.
\newblock Neural machine translation with byte-level subwords.
\newblock In \emph{Proceedings of the AAAI Conference on Artificial
  Intelligence}, volume~34, pages 9154--9160.

\bibitem[{Wang et~al.(2020{\natexlab{b}})Wang, Lipton, and
  Tsvetkov}]{wang-etal-2020-negative}
Zirui Wang, Zachary~C. Lipton, and Yulia Tsvetkov. 2020{\natexlab{b}}.
\newblock \href {https://www.aclweb.org/anthology/2020.emnlp-main.359} {{On
  Negative Interference in Multilingual Models: Findings and A Meta-Learning
  Treatment}}.
\newblock In \emph{Proceedings of the 2020 Conference on Empirical Methods in
  Natural Language Processing (EMNLP)}, pages 4438--4450, Online. Association
  for Computational Linguistics.

\bibitem[{Wang et~al.(2019)Wang, Xie, Xu, Yang, Neubig, and
  Carbonell}]{DBLP:journals/corr/abs-1910-04708}
Zirui Wang, Jiateng Xie, Ruochen Xu, Yiming Yang, Graham Neubig, and Jaime~G.
  Carbonell. 2019.
\newblock \href {http://arxiv.org/abs/1910.04708} {Cross-lingual alignment vs
  joint training: {A} comparative study and {A} simple unified framework}.
\newblock \emph{CoRR}, abs/1910.04708.

\bibitem[{Wolf et~al.(2020)Wolf, Debut, Sanh, Chaumond, Delangue, Moi, Cistac,
  Rault, Louf, Funtowicz, Davison, Shleifer, von Platen, Ma, Jernite, Plu, Xu,
  Scao, Gugger, Drame, Lhoest, and Rush}]{wolf-etal-2020-transformers}
Thomas Wolf, Lysandre Debut, Victor Sanh, Julien Chaumond, Clement Delangue,
  Anthony Moi, Pierric Cistac, Tim Rault, Rémi Louf, Morgan Funtowicz, Joe
  Davison, Sam Shleifer, Patrick von Platen, Clara Ma, Yacine Jernite, Julien
  Plu, Canwen Xu, Teven~Le Scao, Sylvain Gugger, Mariama Drame, Quentin Lhoest,
  and Alexander~M. Rush. 2020.
\newblock \href {https://www.aclweb.org/anthology/2020.emnlp-demos.6}
  {{Transformers: State-of-the-Art Natural Language Processing}}.
\newblock In \emph{Proceedings of the 2020 Conference on Empirical Methods in
  Natural Language Processing: System Demonstrations}, pages 38--45, Online.
  Association for Computational Linguistics.

\bibitem[{Wu and Dredze(2019)}]{wu-dredze-2019-beto}
Shijie Wu and Mark Dredze. 2019.
\newblock \href {https://doi.org/10.18653/v1/D19-1077} {Beto, bentz, becas: The
  surprising cross-lingual effectiveness of {BERT}}.
\newblock In \emph{Proceedings of the 2019 Conference on Empirical Methods in
  Natural Language Processing and the 9th International Joint Conference on
  Natural Language Processing (EMNLP-IJCNLP)}, pages 833--844, Hong Kong,
  China. Association for Computational Linguistics.

\bibitem[{Yang et~al.(2019)Yang, Dai, Yang, Carbonell, Salakhutdinov, and
  Le}]{yang2019xlnet}
Zhilin Yang, Zihang Dai, Yiming Yang, Jaime Carbonell, Russ~R Salakhutdinov,
  and Quoc~V Le. 2019.
\newblock Xlnet: Generalized autoregressive pretraining for language
  understanding.
\newblock In \emph{Advances in neural information processing systems}, pages
  5753--5763.

\bibitem[{Zeman et~al.(2020)Zeman, Nivre, Abrams, and Ackermann~\emph{et
  al.}}]{11234/1-3424}
Daniel Zeman, Joakim Nivre, Mitchell Abrams, and Elia Ackermann~\emph{et al.}
  2020.
\newblock \href {http://hdl.handle.net/11234/1-3424} {Universal dependencies
  2.7}.
\newblock {LINDAT}/{CLARIAH}-{CZ} digital library at the Institute of Formal
  and Applied Linguistics ({{\'U}FAL}), Faculty of Mathematics and Physics,
  Charles University.

\end{thebibliography}
\bibliographystyle{acl_natbib}

\end{document}